%% file: ICML_main.tex
\documentclass{article}

\usepackage{microtype}
\usepackage{graphicx}
\usepackage{subfigure}
\usepackage{booktabs} 

\usepackage[utf8]{inputenc} 
\usepackage[T1]{fontenc}    
\usepackage{hyperref}       
\usepackage{url}            
\usepackage{booktabs}       
\usepackage{amsfonts}       
\usepackage{nicefrac}       
\usepackage{microtype}      
\usepackage{graphicx}
\usepackage{xcolor}
\usepackage{cite}
\usepackage{comment}
\usepackage{wrapfig,lipsum,booktabs}
\usepackage{placeins}
\usepackage{amsmath}
\usepackage[ruled,vlined]{algorithm2e}
\usepackage{float}
\restylefloat{table}
\usepackage{textcomp}

\usepackage{hyperref}

\usepackage[accepted]{icml2021}

\title{Improving Robustness of Learning-based Autonomous Steering \\ Using Adversarial Images}
\icmltitlerunning{Improving Robustness of Learning-based Autonomous Steering}

\DeclareMathOperator*{\minimize}{min}
\DeclareMathOperator*{\maximize}{max}

\DeclareMathOperator*{\argmin}{arg\,min}

\begin{document}

\twocolumn[
\icmltitle{Improving Robustness of Learning-based Autonomous Steering Using Adversarial Images}



\icmlsetsymbol{equal}{*}

\begin{icmlauthorlist}
\icmlauthor{Yu Shen}{UMD}
\icmlauthor{Laura Zheng}{UMD}
\icmlauthor{Manli Shu}{UMD}
\icmlauthor{Weizi Li}{UofM}
\icmlauthor{Tom Goldstein}{UMD}
\icmlauthor{Ming C. Lin}{UMD}
\end{icmlauthorlist}

\icmlaffiliation{UMD}{Department of Computer Science, University of Maryland, Maryland, USA}
\icmlaffiliation{UofM}{Department of Computer Science, University of Memphis, Tennessee, USA}

\icmlcorrespondingauthor{}{\{yushen,lyzheng,manlis,tomg,lin\}@umd.edu}

\icmlkeywords{Robustness, Computer Vision, Deep Learning, ICML}

\vskip 0.3in
]



\printAffiliationsAndNotice{\icmlEqualContribution} 

\newcommand{\weizi}[1]{{\textcolor{orange}{[#1]}}}

\newcommand{\modified}[1]{{\textcolor{black}{#1}}}

\newcommand{\train}{${\rm{train}}$}
\newcommand{\ptrain}{${\rm{ptrain}}$}
\newcommand{\pretrainTrain}[2]{\train$($#1$) \rightarrow $ \train$($#2$)$}
\newcommand{\pretrainPartialRetrain}[2]{\train$($#1$) \rightarrow$ \ptrain$($#2$)$}

\newcommand{\BN}{${\rm{BN}}$}
\newcommand{\MA}{MA}
\newcommand{\MAR}{${\rm{MA}_{R}}$}
\newcommand{\MARM}{${\rm{MA}}_{R}(M)$}
\newcommand{\MARD}[1]{${\rm{MA}}_{R}($\train$($#1$))$}
\newcommand{\MAMD}[2]{${\rm{MA}}_{#2}($#1$)$}
\newcommand{\MADD}[2]{${\rm{MA}}_{#2}($\train$($#1$))$}
\newcommand{\BNMD}[2]{${\rm{BN}}_{#2}($#1$)$}
\newcommand{\BNDD}[2]{${\rm{BN}}_{#2}($\train$($#1$))$}
\newcommand{\BNDDC}[3]{${\rm{BN}}_{#2}^{#3}($\train$($#1$))$}
\newcommand{\BNMDC}[3]{${\rm{BN}}_{#2}^{#3}($#1$)$}
\newcommand{\rBase}{$R$}
\newcommand{\rBasen}{R}
\newcommand{\vBase}{$V$}
\newcommand{\vBasen}{V}
\newcommand{\vrgan}{$T_{C}$}
\newcommand{\vrgann}{T_{C}}
\newcommand{\vrmunit}{$T_{M}$}
\newcommand{\vrmunitn}{T_{M}}
\newcommand{\bOne}{Blur1}
\newcommand{\bTwo}{Blur2}
\newcommand{\bThree}{Blur3}
\newcommand{\bFour}{Blur4}
\newcommand{\bFive}{Blur5}
\newcommand{\nOne}{Noise1}
\newcommand{\nTwo}{Noise2}
\newcommand{\nThree}{Noise3}
\newcommand{\nFour}{Noise4}
\newcommand{\nFive}{Noise5}
\newcommand{\dOne}{Distortion1}
\newcommand{\dTwo}{Distortion2}
\newcommand{\dThree}{Distortion3}
\newcommand{\dFour}{Distortion4}
\newcommand{\dFive}{Distortion5}

\newcommand{\ltwo}{$\mathcal{L}2$}

\begin{abstract}
For safety of autonomous driving, vehicles need to be able to drive under various lighting, weather, and visibility conditions in different environments. These external and environmental factors, along with internal factors associated with sensors, can pose significant challenges to perceptual data processing, hence affecting the decision-making and control of the vehicle. In this work, we address this critical issue by introducing a framework for analyzing robustness of the learning algorithm w.r.t varying quality in the image input for autonomous driving. Using the results of sensitivity analysis, we further propose an algorithm to improve the overall performance of the task of ``learning to steer''. The results show that our approach is able to enhance the learning outcomes up to 48\%. 
A comparative study drawn between our approach and other related techniques, such as data augmentation and adversarial training, confirms the effectiveness of our algorithm as a way to improve the robustness and generalization of neural network training for autonomous driving. 
\end{abstract}

\input{sections/introduction}

\input{sections/related-work}

\input{sections/analysis}

\input{sections/learning-method}
\input{sections/learning-results}
\input{sections/benchmark}
\input{sections/conclusion}

\clearpage

\nocite{langley00}

\bibliography{references}
\bibliographystyle{icml2021}


\input{sections/appendix}

\end{document}

%% file: sections/introduction.tex
\section{Introduction}
Autonomous driving is a complex task that requires many software and hardware components to operate reliably under highly disparate and often unpredictable conditions. 
%
\modified{Steering, as an end-to-end autonomous driving task, contains both perception and control (two of the most important components in autonomous driving systems), which makes it an ideal target task to explore.}
While on the road, vehicles are likely to experience 
day and night, clear and foggy conditions, sunny and rainy days, as well as bright cityscapes and dark tunnels. All these external factors in conjunction with internal factors of the camera (e.g., those associated with hardware) can lead to quality variations in image data, which are then served as input to image-based learning algorithms. 
\modified{One can harden machine learning systems to these degradations by simulating them at training time~\citep{Chao2019Survey}.}
However, there currently lacks algorithmic tools for analyzing the sensitivity of real-world neural network performance on the properties of training images and, more importantly, a mechanism to leverage such a sensitivity analysis for improving learning outcomes. In this work, we quantify the influence of image quality on the task of ``learning to steer,'' study how training on degraded and low-quality images can boost robustness to image corruptions, and provide a systematic approach to improve the performance of the learning algorithm based on quantitative analysis.  

Image degradations can often be simulated at training time by adjusting a combination of image quality attributes, including blur, noise, distortion, color representations (such as RGB or CMY) hues, saturation, and intensity values (HSV), etc.  However, identifying the correct combination of the simulated attribute parameters to obtain optimal performance on real data during training is a difficult---if not impossible---task, as it requires domain transfer and exploring a high dimensional parameterized space.

The first goal of this work is to \emph{design a systematic method for measuring the severity of an image degradation, and predicting the impact it will have on system performance}. \modified{Inspired by the use of variance in image features in sensitivity analysis for ML models~\cite{saltelli2008global}}, we choose to measure the difference between real-world image distributions and simulated/degraded image distributions using the Fr\'{e}chet Inception Distance (FID). 
\modified{Our experimental results confirm that the FID between simulated and real datasets helps predict the performance of systems trained using simulated data and deployed in the real world.}
We also use FID between different simulated datasets as a unified metric to parameterize the severity of various image degradations (Section~\ref{sec:quality_analysis}).

Our second goal is to borrow concepts from the adversarial attack literature~\citep{madry2017towards, shafahi2019adversarial, xie2019adversarial} to \emph{build a scalable training scheme for enhancing the robustness of autonomous driving systems against multi-faceted image degradations, while \emph{increasing} the overall accuracy of the steering task on clean data}. 
Our proposed method builds a dataset of adversarially degraded images by apply evolutionary optimization within the space of possible degredations during training. The method begins by training on a combination of real and simulated/degraded images using arbitrary degradation parameters. On each training iteration, the parameters are updated to generate a new degradation combination so that the system performance is (approximately) minimized.  The network is then trained on these adversarially degraded images to promote robustness. Our proposed algorithm uses our FID-based parameterization to discretize the search space of degradation parameters and accelerateå the process of finding optimal parameters  (Section~\ref{sec:rob_train}). 

Experiments show that our algorithm improves task performance for ``learning to steer'' up to \textbf{48\%} in mean accuracy over strong baselines. We compare our approach with other tasks (e.g., detection) and related techniques, such as data augmentation and adversarial training. The results show that our method consistently achieves higher performance. Our technique also improves the performance on datasets contaminated with complex combinations of perturbations (up to \textbf{33\%}), and additionally boosts the test performance on degradations that are not seen during training, including simulated snow, fog, and frost (up to \textbf{29\%}) (Section~\ref{results}).





For evaluation, we propose a more comprehensive robustness evaluation standard under four different scenarios: clean data, single-perturbation data, multi-perturbation data, and previously unseen data. While state-of-the-art studies usually conduct testing under one or two scenarios, our work is among the first to test and verify results under four meaningful scenarios for evaluating the robustness of a learning algorithm.

We plan to release code and ``autonomous driving under perturbations'' datasets for benchmarking, which will include a base dataset, the simulated adversarial datasets with multiple levels of image degradation using either single or multiple image attributes, and the simulated adversarial datasets with multiple levels of combinatorial perturbations using unseen factors for
image corruptions in ImageNet-C~\citep{hendrycks2019benchmarking}, totaling \textbf{360} datasets and \textbf{2.2~M} images.

%% file: sections/related-work.tex
\begin{figure*}[t]
\begin{center}
  \includegraphics[width=\linewidth]{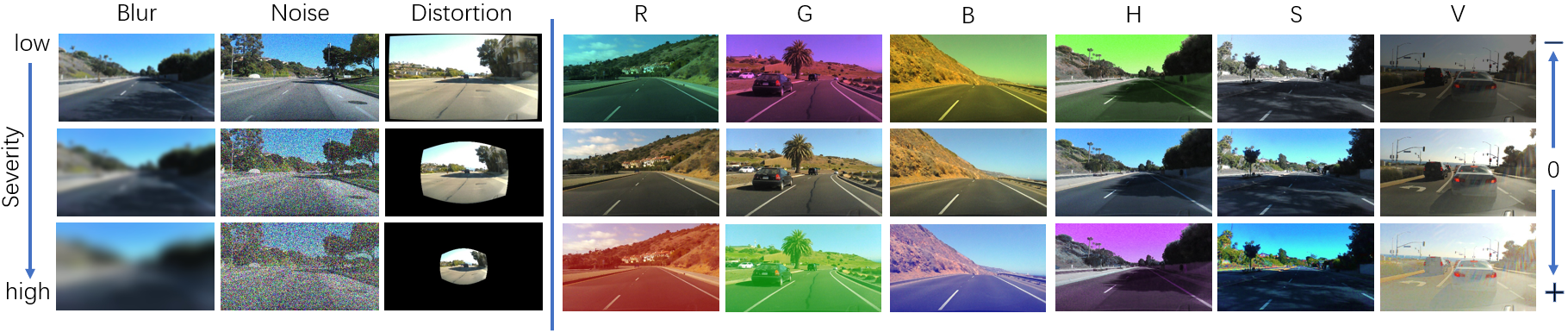}
\end{center}   
\vspace*{-1.75em}
\caption{
\modified{Example images of quality degradation. Left: Five levels of quality reduction are simulated using the blur, noise, and distortion effects (three are shown). Right: The original images are shown in the middle row. The top row shows examples in the lighter direction per channel for R, G, B, H, S, V, while the bottom row shows examples in the darker direction per channel.}}
\label{fig:driving_image_quality_all}
\vspace*{-0.75em}
\end{figure*}

\section{Related Work}

The influence of the noise and distortion effects on real images for learning tasks has been well explored. For example, researchers have examined the impact of optical blur on convolutional neural networks and present a fine-tuning method for recovering lost accuracy using blurred images~\citep{blur-impact}. 
This fine-tuning method resolves lost accuracy when images are distorted instead of blurred~\citep{distorted-impact}. While these fine tuning methods are promising,~\citep{quality_resilient} find that tuning to one type of image quality reduction effect would cause poor generalization to other types of quality reduction effects.
The comparison of image classification performance between deep neural networks and humans is conducted~\citep{noise-distortion}, and found to be similar with images of good quality.  However, deep neural networks struggle significantly more than humans on low-quality, distorted, and noisy images. Color spaces have also been shown to negatively affect the performance of learning algorithms. 
One study shows that adversarial perturbations are more prevalent in the Y channel in the YCbCr color space of images than the other two channels, while perturbations in RGB channels are equally distributed~\citep{yuv-y}.
~\citet{instagram-filter} studies the effect of Instagram filters, which mostly change the coloring of an image, on learning tasks. In this work, we study nine common factors characterizing image quality, i.e., blur, noise, distortion, three-color (RGB) channels, and hues, saturation, and intensity values (HSV). Not only does our study analyze a more comprehensive set of image attributes that could influence the learning-based steering task, but we also parameterize these nine factors into one integrated image-quality space using the Fr\'{e}chet Inception Distance as the unifiying metric, thus enabling the sensitivity analysis. 


Researchers have also explored how to improve the robustness of learning algorithms under various image quality degradations. One recent work~\citep{tran2017bayesian} provides a novel Bayesian formulation for data augmentation.~\citet{cubuk2018autoaugment} proposes an approach to automatically search for improved data augmentation policies.~\citet{ghosh-robustness} performs analyses on the performance of convolutional neural networks on quality degradations because of causes such as compression loss, noise, blur, and contrast, and introduces a method to improve the learning outcome. 
Another work~\citep{hendrycks-robustness-selfsupervised} shows that self-supervision techniques can be used to improve model robustness and exceeds the performance of fully-supervised methods. 
A new method, also by Hendrycks et al., improves model robustness using data augmentation, where transformation compositions are used to create a new dataset, which is visually and semantically similar to the original dataset~\citep{hendrycks2019augmix}.
~\citet{gao2020fuzz} proposes a technique to re-purpose software testing methods to augment the training data of DNNs, with the objective to improve model robustness. A recent work~\citep{maxup} improves model generalizability by first augmenting training dataset with random perturbations, and then minimizing worst-case loss over the augmented data. 

Our work differs from these studies in several regards. First, we simulate adversarial conditions of image factors instead of using commonplace image conditions. Second, we conduct a systematic sensitivity analysis for preparing datasets that are representative of image degradations from multiple factors at various levels. Third, our algorithm can work with the discretized parameter space while generalizing well to the continuous parameter space. Another advantage of our approach is that we can augment the training dataset without the derivatives of the factor parameters, which may not exist or are difficult to compute. These differences distinguish our approach to previous studies for improving model robustness. 






%% file: sections/analysis.tex
\section{Analyzing Image Quality Factors}
\label{sec:quality_analysis}

\subsection{Experiment Setup}
\label{sec:experiment_setup}

\modified{\textbf{Base Dataset}. Our base dataset consists of three real-world driving datasets:  Audi~\citep{geyer2020a2d2}, Honda~\citep{ramanishka2018toward}, and SullyChen~\citep{NVIDIA-Data}. Among the three, the Audi dataset is the most recent (2020), the Honda dataset has many driving videos (100+), and the SullyChen dataset focuses on the steering task and has the longest continuous driving image sequence without road branching. The details of these datasets can be found in Appendix~\ref{Apd:dataset_details}.}

\textbf{Image quality factors}. We study nine image attributes in this work: blur, noise, distortion, three-color (RGB) channels, and hues, saturation, and intensity values (HSV). Blur, noise, and distortion are among the most commonly used factors that can directly affect the image quality. 
R, G, B, H, S, V channels are chosen because both RGB and HSV are frequently used to represent image color spaces: RGB represent three basic color values of an image, while HSV represent three common metrics of an image. Other color spaces such as HSL or YUV have similar properties, hence are excluded from this study. 

\textbf{Learning Algorithm}. We choose the model by~\citet{bojarski2016end} as the learning algorithm. The model contains five convolutional layers followed by three dense layers. We select this model, because it has been used to steer an AV successfully in both real world~\citep{bojarski2016end} and virtual world~\citep{Li2019ADAPS}.

\begin{figure*}[t]
\begin{center}
  \includegraphics[width=0.9\linewidth]{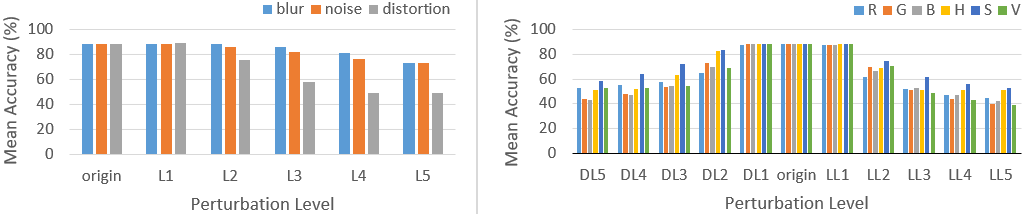}
\end{center}   
\vspace*{-1.5em}
\caption{The relationship between MA and levels of perturbation: greater image degradations lead to higher loss of mean accuracy.}
\label{fig:MA_quality_all}
\vspace*{-1em}
\end{figure*}

\textbf{Evaluation Metric}. We use mean accuracy (\MA{}) to evaluate the learning task since it can represent the overall performance under a variety of thresholds (similar to Mean Average Precision (mAP) for the detection task). We first define the accuracy with respect to a particular threshold $\tau$ as $acc_{\tau}=count(|v_{predicted} - v_{actual}|<\tau)/n$, where $n$ denotes the number of test cases; $v_{predicted}$ and $v_{actual}$ indicate the predicted and ground-truth value, respectively. Then, we compute \MA{} as $\sum_{\tau}acc_{\tau \in \mathcal{T}}/|\mathcal{T}|$, where 
$\mathcal{T}=\{1.5, 3.0, 7.5, 15, 30, 75\}$ contains empirically selected thresholds of steering angles. Lastly, we use the maximum \MA{} improvement (denoted as MMAI), the average \MA{} improvement (denoted as AMAI), and mean Corruption Errors (mCE)~\citep{hendrycks2019benchmarking} as the evaluation metrics. 


\subsection{Sensitivity Analysis}
\label{sec:ssensitivity_analysis}

\modified{Sensitivity analysis (SA) is commonly used to understand how the uncertainty in the model output (numeric or otherwise) can be apportioned to different sources of uncertainty in the model input~\citep{saltelli2002sensitivity}~\citep{saltelli2008global}. 
A recent work~\citep{zhang2015sensitivity} shows SA can be used to better interpret CNNs. There are a range of purposes of SA, e.g., testing the robustness of model results in the presence of uncertainty, understanding the relationship between model input and output variables, etc. Here, we use SA to study how distortions to model inputs (blur, noise, distortion and RGB/HSV brightness shifts) can be apportioned to model output (i.e., \MA{} of the predicted steering angle for the test data). We also use SA results to prepare datasets that are representative of image qualities at various degradation levels for model training.
}

\modified{Specifically, we focus on the changes in model performance according to the changes in the input factors: $sensitivity = \frac{\partial E(Y)}{\partial X}$, where $Y$ is a random variable representing model performance, while $X$ represents the input factor(s). Note that $Y$ is likely to have confounding variables beyond $X$ due to optimization stochasticity. However, as our focus is neither the stability of the optimizer nor the training process of neural network, we simplify our objective to $$sensitivity = \frac{\partial Y}{\partial X},$$ under the assumption that the optimizer and training process are stable given the same data and configurations. 
}

Next, we introduce the details of our analysis. For simulating blur, noise, and distortion effects, we use Gaussian blur~\citep{begin2004blind} (w.r.t standard deviation), additive white Gaussian noise (AWGN) (w.r.t standard deviation), and radial distortion~\citep{zhang2000flexible} (w.r.t radial distortion parameter $k1, k2$), respectively. For representing channel-level perturbations, we use a linear model: denote the value range of one channel as $[a,b]$, in the darker direction, we set $v_{new}=\alpha a + (1-\alpha) v$; in the lighter direction, we set $v_{new}=\alpha b + (1-\alpha) v$. The default values are $a_C=0$ and $b_C=255$, where $C$ represents one channel. To exclude a complete dark image, we set $a_V=10$ and $b_H=179$.

We adopt the Fr\'{e}chet Inception Distance (FID)~\citep{heusel2017gans} as a unified metric for our sensitivity analysis (instead of using the parameter values of each image factor) for three reasons. First, 
\modified{given the autonomous driving system is nonlinear, variance-based measures would be more effective for sensitivity analysis of the network.} FID can better capture different levels of image qualities than the parameter values of each factor, because the correspondence between the parameter values and image quality of each factor is not linear.  For example, if we uniformly sample the parameter space of distortion, most values will result in similar images to those of the level 4 and level 5 shown in Fig.~\ref{fig:driving_image_quality_all}. This is problematic as we need representative datasets to better reflect the \emph{sensitivity} of a learning task to an image attribute. Second, using FID, we can map the parameter spaces of all factors into one space to facilitate the sensitivity analysis. Lastly, FID serves as a comprehensive metric to evaluate the distance between two image datasets: image pixels and features, and correlations among images---these meaningful factors to interpret the performance of a learning-based task---are all taken into consideration. \modified{Compared to metrics that only consider image pixels (e.g., L2 norm distance), FID can better distinguish the effects due to perturbations (comparisons can be found in Appendix~\ref{Apd:FID_L2D}).} 




\modified{We define the sensitivity as the first-order derivative of \MA{} w.r.t FID:}
\vspace*{-1em}
\modified{
$$sensitivity = \frac{\partial MA^{*}(R \bigoplus F(\mathbf{p}))}{\partial FID(R, R \bigoplus F(\mathbf{p}))},$$
\noindent
where $MA^{*}(D)$ is the \MA{} test result on dataset $D$ with the model trained on the base dataset $R$, $FID(A, B)$ is the FID between datasets $A$ and $B$, $F(\mathbf{p})$ is the perturbation with parameter $\mathbf{p}$ (e.g., the standard deviation of the Gaussian kernal), and $D \bigoplus F$ means the perturbed dataset by applying perturbation $F$ to $D$.}

Starting from empirically-selected parameters of each factor, we generate perturbed datasets and compute their corresponding \MA{} using the trained model on $R$. We then map the correspondences between the MAs and the parameter values into the FID space. 
\modified{By leveraging this new MA-FID space, we can reduce and minimize the number of parameter samples for each factor, while the sampled dataset provides a similar curve to the original one to improve computational efficiency of training (see Section~\ref{sec:rob_train}).}
Examples of the resulting images are shown in Fig.~\ref{fig:driving_image_quality_all} (more can be seen in Appendix~\ref{Apd:image_samples}). Detailed descriptions of the final perturbed datasets are provided in Appendix~\ref{Apd:perturbed-datasets}.

The analysis results using the final perturbed datasets are shown in Fig.~\ref{fig:MA_quality_all}. The complete numeric results can be found in Appendix~\ref{Apd:experiment_data}.  Note that all factors have some influence on the learning-based steering task. As the perturbation level increases, their (negative) impact increases. The performance loss because of image quality degradation can be higher than 50\% (see lighter Lv5 of the G channel), which can impose significant risk for autonomous driving.

\begin{figure}[t]
\begin{center}
  \includegraphics[width=\linewidth]{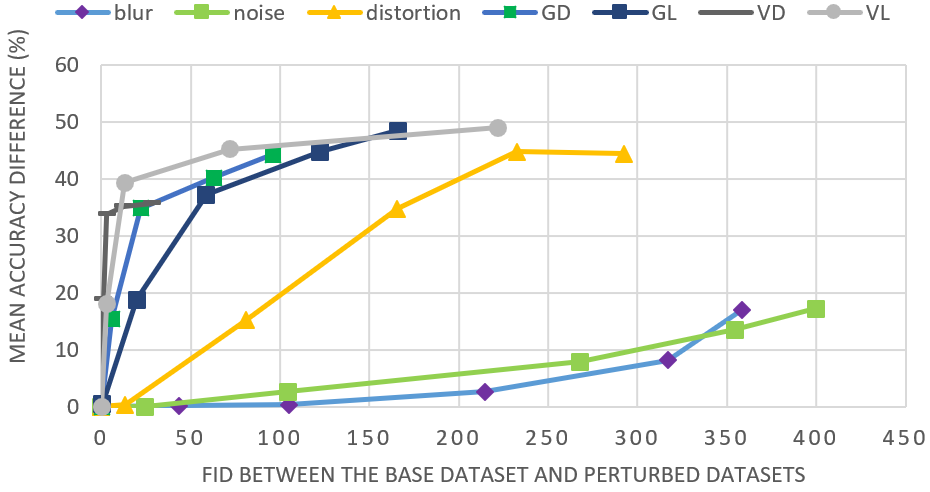}
\end{center}   
\vspace*{-1.5em}
\caption{The relationship between FID and MA difference. GD/GL denotes G channel in darker/lighter direction, and VD/VL denotes V channel in darker/lighter direction, respectively. Sensitivity is represented by the first-order derivative of the curve.}
\label{fig:FID_MAdiff}
\vspace*{-1em}
\end{figure}

\begin{figure*}[t]
\begin{center}
\includegraphics[width=\linewidth]{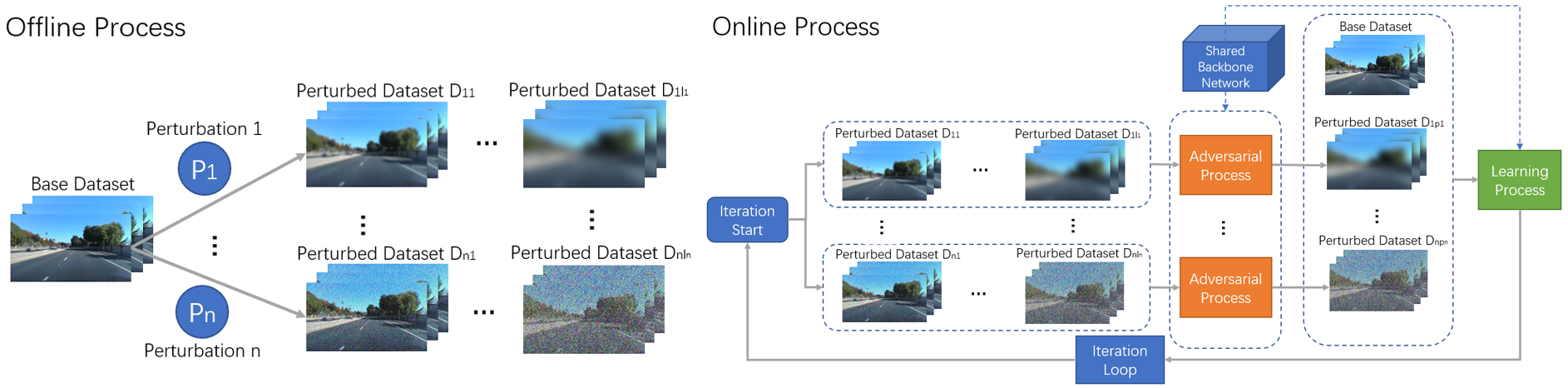}
\end{center}   
\vspace*{-1.5em}
\caption{
\modified{Pipeline of our method. Offline: we generate perturbed datasets of each factor at multiple levels based on the sensitivity analysis results. Online: in each iteration, we first augment the training dataset with ``adversarial images'' given each perturbation, i.e., select the datasets with the worst performance of each perturbation, then combine the base dataset and those to train our model to maximize the overall performance.}}
\label{fig:overall_pipeline}
\vspace*{-1em}
\end{figure*}

The final MA differences in the FID space are visualized in Fig.~\ref{fig:FID_MAdiff} (for blur, noise, distortion, and G and V V channels; see entire figure in Appendix~\ref{Apd:FID_L2D}). Since FID aligns different factors into the same space, we can compare the performance loss of all factors at various levels. Notice that the values in the near-zero FID range (i.e., FID$<50$) are more commonly found in real-world applications. We first observe that the learning-based steering task is more sensitive to the channel-level perturbations (i.e., R, G, B, H, S, V) than the image-level perturbations (i.e., blur, noise, distortion). Second, the task is least sensitive to blur and noise but most sensitive to the V channel, the intensity value. Third, for the same color channel, darker and lighter levels appear to have different MAs at the same FID values. Compared with other analysis studies on the ``learning to steer'' task, e.g.,~\citet{tian2018deeptest} and~\citet{zhang2018deeproad}, our method is the first to transfer perturbations from multiple parameter spaces into one unified space to enable the cross-factor comparison, e.g., the task is more sensitive to the V-channel perturbation than perturbations in other attributes.

   
   

%% file: sections/learning-method.tex
\vspace*{-0.5em}
\section{Robustness of Learning-based Steering}
\label{sec:rob_train}

\subsection{Methodology}
In this section, we introduce our method to improve the generalization of learning-based steering using the acquired sensitivity analysis results.

Our algorithm uses an iterative min-max training process: at each iteration, we first choose a dataset from all datasets of one factor that can minimize \MA{}. Then, we combine such datasets of all factors with the base dataset to re-train our model while maximizing MA. The algorithm stops when a pre-specified number of iterations is reached or the MA loss is below a certain threshold. The design rationale of our architecture resembles adversarial training: we train the model to maximize accuracy using the original dataset plus the perturbed datasets with the minimum accuracy in order to improve the robustness of the model. The loss function is the following:
\begin{equation*}
   \minimize_{\mathbf{p}} \maximize_{\theta} MA(\theta, U_{\mathbf{p}}),
   \vspace*{-1em}
\end{equation*}

where $\mathbf{p}$ represents a union of the parameter levels of all analyzed image factors, $\theta$ denotes the model parameters, $U_{\mathbf{p}}$ is the training dataset, and $\rm{MA}()$ is the function computing MA. Our method is described in Algorithm~\ref{alg:robustness_train}; the pipeline is shown in Fig.~\ref{fig:overall_pipeline}.

\begin{algorithm}[th]
\label{alg:robustness_train}
 \caption{
 Improve robustness of learning-based steering
 }
\begin{algorithmic}
\SetAlgoLined
\STATE {\bfseries Result:} a trained model parameterized by $\theta$

\STATE {\bfseries Pre-processing:} 
 
 Conduct sensitivity analysis and discretize the parameters of $n$ factors into their corresponding $l_i, i=1,\dots,n$ levels
 
 Generate new datasets for each factor with the discretized values from the base dataset $R$: $\{D_{i,j}\}, i=1,2,..,n, j=1,2,..l_i$
 
\STATE {\bfseries  Initialization: } 

 Set $t=0$, and initialize the maximum iterations $T$ and the number of epochs $k$
 
 Initialize model parameters $\theta^{(0)}$
 
\STATE {\bfseries  Iteration: }

 \While{$t \le T$}{
  For each factor, select the dataset $D_{i,p_i}$ that can minimize the validation \MA{}, where
  $p_i=\argmin_j MA(\theta^{(t)}, D_{ij})$
  
  Merge all selected datasets 
  $U_{\mathbf{p}}=(\bigcup_{i=1}^{n} D_{i,p_i}) \bigcup R$
  
  Train the network for $k$ epochs and update 
  $\theta^{(t+1)}$ = train($\theta^{(t)}$, $U_{\mathbf{p}}$, $k$)
  to maximize $MA(\theta^{(t+1)}, U_{\mathbf{p}})$
  
  Break if stop conditions are met; otherwise set $t=t+1$
 }
\end{algorithmic}
\end{algorithm}

\begin{table*}[th]
  \centering
  \scalebox{.8}{
  \begin{tabular}{c|c|ccc|ccc|ccc}
    \toprule
    & \multicolumn{10}{c}{Scenarios} \\
        \midrule
      & \multicolumn{1}{c}{Clean} & \multicolumn{3}{c}{Single Perturbation} & \multicolumn{3}{c}{Combined Perturbation} & \multicolumn{3}{c}{Unseen Perturbation} \\
    \midrule
   Method& AMAI$\uparrow$ & MMAI$\uparrow$ & AMAI$\uparrow$ & mCE$\downarrow$ & MMAI$\uparrow$ & AMAI$\uparrow$ & mCE$\downarrow$ & MMAI$\uparrow$ & AMAI$\uparrow$ & mCE$\downarrow$\\
    \midrule
    
Data Augmentation & -0.44 & 46.88 & 19.97 & 51.34 & \textbf{36.1}  & 11.97 & 75.84 & 27.5  & 7.92 & 81.51 \\
Adversarial Training & -0.65 & 30.06 & 10.61 & 74.42 & 17.89  & 6.99 & 86.82 & 16.9  & 8.17 & 89.91\\
MaxUp & -7.79 & 38.30 & 12.83 & 66.56 & 26.94 & 16.01 & 72.60 & 23.43 & 5.54 & 81.75\\
AugMix & -5.23 & 40.27 & 15.01 & 67.49 & 26.81 & 15.45 & 68.38 & 28.70 & 8.85 & 87.79\\
Ours & \textbf{0.93} & \textbf{48.57} & \textbf{20.74} & \textbf{49.47} & 33.24  & \textbf{17.74} & \textbf{63.81} & \textbf{29.32}  & \textbf{9.06} & \textbf{76.20} \\
    \bottomrule
  \end{tabular}}
 \caption{
\modified{Performance of different methods against the baseline performance~\citep{bojarski2016end} on SullyChen dataset. The evaluation metrics are the maximum MA improvement in percentage (MMAI), the average MA improvement in percentage (AMAI), and mean corruption errors in percentage (mCE). We compare basic data augmentation method (simply combine all perturbed datasets into training), an adversarial training method~\citep{shu2020preparing}, MaxUp~\citep{maxup}, and AugMix~\citep{hendrycks2019augmix}. Overall, our method outperforms all other methods (i.e., highest MA improvements and lowest error in mCEs) in practically all scenarios.}}
  \label{tb:comparison_methods}
\end{table*}



Our method offers several advantages: 1) the training data is augmented without re-train the model, thus improving efficiency; 2) it provides the flexibility to generate datasets at various discretized levels of the factor parameters; 3) it does not require the derivatives of factor parameters (other methods that optimize factor parameters in the continuous space require computing derivatives), which could be difficult; and 4) it can generalize to not only unseen parameters of individual factors but also the composition of unseen parameters of multiple factors.

%% file: sections/learning-results.tex
\subsection{Experiments and Results}
\vspace*{-0.5em} 
\label{results}
\textbf{Setups:} All experiments are conducted using Intel(R) Xeon(TM) W-2123 CPU, Nvidia GTX 1080 GPU, and 32G RAM. We use the Adam optimizer~\citep{kingma2014adam} with learning rate 0.0001 and batch size 128 for training. The maximum number of epochs is 4,000. The datasets setup is shared with analysis experiments (see Section~\ref{sec:experiment_setup}).

\textbf{Test scenarios and metrics:} We test the performance of all methods in four scenarios with increasing complexity.

Scenario 1: \emph{Clean data}. Test on the base clean dataset only.

Scenario 2: \emph{Single Perturbation}. Test on the datasets of each factor at their corresponding discretized levels. Specifically, we use five levels for blur, noise, distortion, and ten levels for R, G, B, H, S, V. In total, there are 75 datasets. 

Scenario 3: \emph{Combined Perturbation}. Test on the datasets with combinations of all factors at all levels. To be specific, we sample varying levels of each factor, and combine the resulting datasets of all factors into one \emph{combined} dataset. In total, we have six \emph{combined} datasets. See examples in the second row of Figure~\ref{fig:salience_map}. The details of these datasets are provided in Appendix~\ref{Apd:perturbed-datasets}.

\begin{figure}[h]
\vspace*{-0.75em}
\begin{center}
  \includegraphics[width=\linewidth]{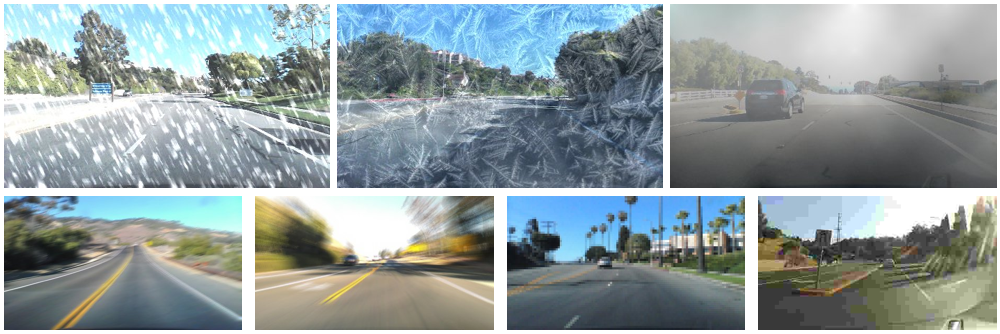}
\end{center}   
\vspace*{-1.5em}
\caption{Unseen perturbation examples in our experiments. We use ``snow'', ``frost'', ``fog'' (left to right; first row), and ``motion blur'', ``zoom blur'', ``pixelate'', ``jpeg compression'' (left to right; second row) from the corruptions in ImageNet-C~\citep{hendrycks2019benchmarking}. 
}
\label{fig:unseen_factors}
\vspace*{-0.75em}
\end{figure}

Scenario 4: \emph{Unseen Perturbation}. Test on the datasets using previously unseen factors at different levels. The unseen factors are ``motion blur'', ``zoom blur'', ``pixelate'', ``jpeg compression'', ``snow'', ``frost'', and ``fog'' from ImageNet-C~\citep{hendrycks2019benchmarking}. We choose these factors because ``Motion blur'' and ``zoom blur'' can happen during driving; ``pixelate'' and ``jpeg compression'' are possible during image processing; and ``snow'', ``frost'', ``fog'' are natural conditions, which can affect the driving experience
(see examples in Figure~\ref{fig:unseen_factors}).

\begin{table*}[t]
  \centering
  \scalebox{.8}{ 
  \begin{tabular}{c|c|ccc|ccc|ccc}
    \toprule
    & \multicolumn{10}{c}{Scenarios} \\
        \midrule
      & \multicolumn{1}{c}{Clean} & \multicolumn{3}{c}{Single Perturbation} & \multicolumn{3}{c}{Combined Perturbation} & \multicolumn{3}{c}{Unseen Perturbation} \\
    \midrule
   Method& AMAI$\uparrow$ & MMAI$\uparrow$ & AMAI$\uparrow$ & mCE$\downarrow$ & MMAI$\uparrow$ & AMAI$\uparrow$ & mCE$\downarrow$ & MMAI$\uparrow$ & AMAI$\uparrow$ & mCE$\downarrow$\\
    \midrule
    
AugMix+Nvidia & -0.12 & 40.64 & 10.94 & 76.48 & 25.97 & 16.79 & 64.41 & \textbf{22.23} & 5.99 & 84.95\\
Ours+Nvidia & \textbf{2.48} & \textbf{43.51} & \textbf{13.51} & \textbf{67.78} & \textbf{28.13} & \textbf{17.98} & \textbf{61.12} & 16.93 & \textbf{6.70} & \textbf{80.92}\\
    \midrule
AugMix+Comma.ai & -5.25 & 55.59 & 9.56 & 86.31 & 31.32 & \textbf{0.77} & \textbf{106.1} & 37.91 & 7.97 & 89.99\\
Ours+Comma.ai & \textbf{0.36} & \textbf{62.07} & \textbf{15.68} & \textbf{70.84} & \textbf{38.01} & 0.74 & 108.32 & \textbf{42.54} & \textbf{12.15} & \textbf{77.08}\\
    \midrule
AugMix+ResNet152 & -4.23 & 20.84 & 1.45 & 96.24 & 12.21 & 6.71 & 80.19 & 15.40 & 2.87 & 97.62\\
Ours+ResNet152 & \textbf{-0.96} & \textbf{24.29} & \textbf{5.19} & \textbf{79.76} & \textbf{16.05} & \textbf{8.02} & \textbf{75.16} & \textbf{16.58} & \textbf{5.33} & \textbf{85.68}\\
    \bottomrule
  \end{tabular}}
  \vspace*{-0.5em}
 \caption{
\modified{Performance improvement of different backbone networks against the baseline performance using the Honda dataset. Our method outperforms AugMix in most cases. Notice that the methods with ResNet152 do not improve as much as the first two networks because the ResNet152 baseline already has relatively high performance.
}}
  \label{tb:comparison_backbones}
\end{table*}

\begin{table*}[t]
  \centering
  \scalebox{.8} { 
  \begin{tabular}{c|c|ccc|ccc|ccc}
    \toprule
    & \multicolumn{10}{c}{Scenarios} \\
        \midrule
      & \multicolumn{1}{c}{Clean} & \multicolumn{3}{c}{Single Perturbation} & \multicolumn{3}{c}{Combined Perturbation} & \multicolumn{3}{c}{Unseen Perturbation} \\
    \midrule
  Method& AMAI$\uparrow$ & MMAI$\uparrow$ & AMAI$\uparrow$ & mCE$\downarrow$ & MMAI$\uparrow$ & AMAI$\uparrow$ & mCE$\downarrow$ & MMAI$\uparrow$ & AMAI$\uparrow$ & mCE$\downarrow$\\
    \midrule
    
AugMix on SullyChen & -5.23 & 40.27 & 15.01 & 67.49 & 26.81 & 15.45 & 68.38 & 28.70 & 8.85 & 87.79\\
Ours on SullyChen & \textbf{0.93} & \textbf{48.57} & \textbf{20.74} & \textbf{49.47} & \textbf{33.24}  & \textbf{17.74} & \textbf{63.81} & \textbf{29.32}  & \textbf{9.06} & \textbf{76.20}\\
    \midrule
AugMix on Honda & -0.12 & 40.64 & 10.94 & 76.48 & 25.97 & 16.79 & 64.41 & \textbf{22.23} & 5.99 & 84.95\\
Ours on Honda & \textbf{2.48} & \textbf{43.51} & \textbf{13.51} & \textbf{67.78} & \textbf{28.13} & \textbf{17.98} & \textbf{61.12} & 16.93 & \textbf{6.70} & \textbf{80.92}\\
    \midrule
AugMix on Audi & -8.24 & 81.89 & 32.22 & 55.27 & 75.49 & 50.23 & 41.98 & 73.06 & 27.39 & 77.51\\
Ours on Audi & \textbf{4.13} & \textbf{94.95} & \textbf{45.78} & \textbf{18.79} & \textbf{80.42} & \textbf{59.31} & \textbf{29.33} & \textbf{75.16} & \textbf{31.91} & \textbf{42.89}\\
    
    \bottomrule
  \end{tabular}}
  \vspace*{-0.5em}
 \caption{
\modified{Performance improvement of different datasets against the baseline performance using the Nvidia backbone. Our method outperforms AugMix in most cases.}}
  \label{tb:comparison_datasets}
\end{table*}


We use the maximum MA improvement (MMAI), the average MA improvement (AMAI), and mean Corruption Errors (mCE)~\citep{hendrycks2019benchmarking} as the evaluation metrics.

\modified{
\textbf{Comparison with different methods:} We compare our method with four other methods: an adversarial training method~\citep{shu2020preparing}, a basic data augmentation method, MaxUp~\citep{maxup}, and AugMix~\citep{hendrycks2019augmix}, to see the performance improvement over the baseline method~\citep{bojarski2016end}. For the basic data augmentation method, we simply merge all perturbed datasets when training the model. }

\modified{
From Table~\ref{tb:comparison_methods}, we observe that our method outperforms other methods under all metrics in all scenarios: not only on the clean dataset but also on perturbed datasets. Notably, our algorithm improves the performance of ``learning to steer'' up to 48\% in MMAI, while reducing mCE by 50\% over the baseline (Scenario 2). Our method also improves the task performance using the \emph{combined} datasets (Scenario 3) up to 33\%. Lastly, when tested on unseen factors (Scenario 4), our algorithm maintains the best performance by 29\% in MMAI, while reducing mCE to 76\%.
}

\modified{
Compared to AugMix~\citep{hendrycks2019augmix}, our adversarial approach can select the most challenging datasets for training, thus improving model robustness. MaxUp~\citep{maxup} picks only the worst case among all augmentation data, which may lead to the loss of data diversity. 
In contrast, our method selects the worst cases in \emph{all} perturbation types (i.e., one dataset per factor), thus improving generalizability. Compared to~\citep{shu2020preparing}, which performs an adversarial process on the entire pixel space with only norm constraints, our method is able to utilize vast prior information generated by sensitivity analysis and reduce the search space. Lastly, compared to the basic data augmentation method, which uses all generated data in training, our method selects the most useful data for training, and thus improves computational efficiency.
}

\modified{
\textbf{Comparison with different backbones:}
We also perform comparison on three backbones: Nvidia network~\citep{bojarski2016end}, Comma.ai network~\citep{santana2016learning}, and ResNet152~\citep{he2016deep}. We conduct these experiments on the Honda dataset. The results shown in Table~\ref{tb:comparison_backbones} indicate that our method achieves higher improvements than AugMix in most cases. Generally, our method can achieve better performance on shallow networks than deep networks. But even on very deep networks such as ResNet152, our method can achieve more than 5\% improvement in all cases, except Scenario 1. 
}

\modified{
\textbf{Comparison on different datasets:}
To demonstrate that our method does not overfit a particular dataset, we experiment on three independent datasets: Audi~\citep{geyer2020a2d2}, Honda~\citep{ramanishka2018toward}, and SullyChen~\citep{NVIDIA-Data}. We use the Nvidia network as the backbone for these experiments. Table~\ref{tb:comparison_datasets} shows that our method can achieve consistently better performance across all three datasets. Furthermore, our method can obtain up to 95\% improvement in some cases. 
}

\modified{
\textbf{Detailed MA improvements:} We also illustrate the MA improvements in Fig.~\ref{fig:MA_channel_unseen_improvement} of Appendix~\ref{Apd:experiment_data}, which shows that our method can achieve improvements at certain channel-factor levels and some unseen image effects. Our method does not improve on ``motion blur'', and we plan to study this issue in future. More detailed data, results, and analysis can be found in Appendix~\ref{Apd:experiment_data}.
}


\textbf{Effectiveness visualization:} Using the salience map on several combined samples in Fig.~\ref{fig:salience_map}, we show our method can help the network to focus on important areas (e.g., the road in front) instead of random areas on perturbed images.

\begin{figure}[h]
\vspace*{-0.5em}
\begin{center}
  \includegraphics[width=\linewidth]{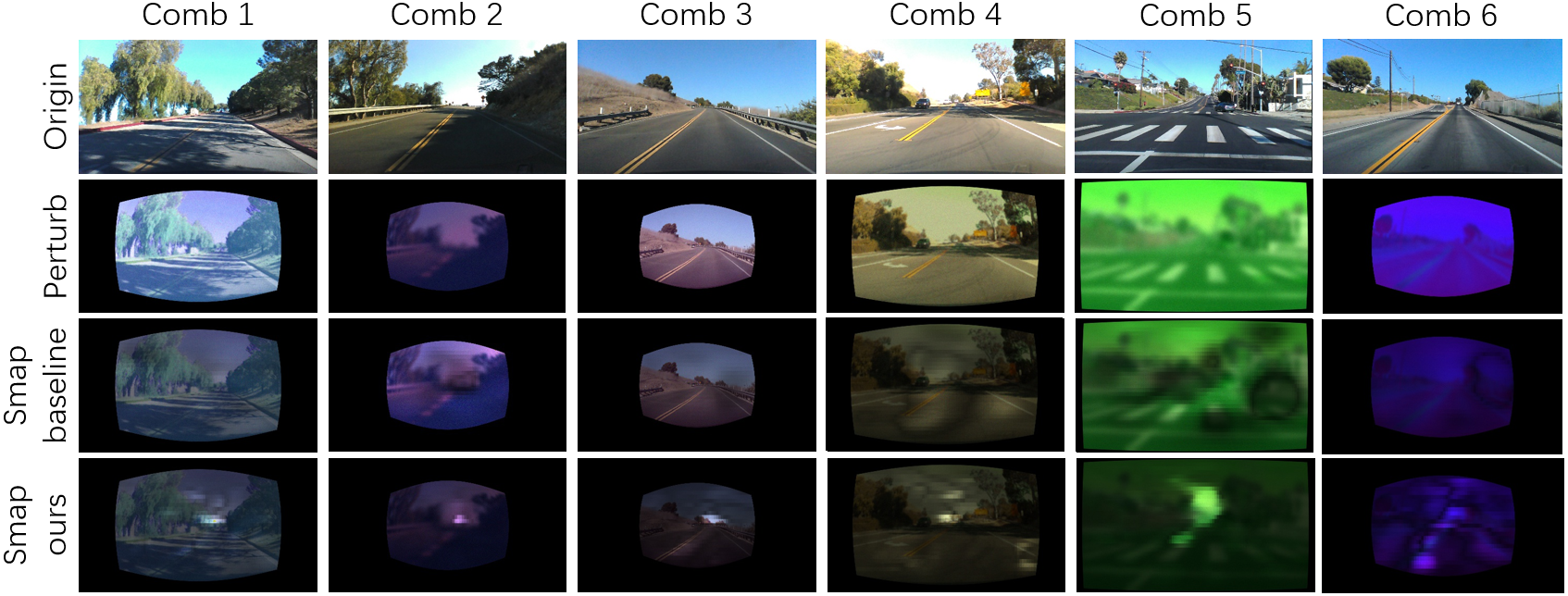}
\end{center}   
\vspace*{-1.5em}
\caption{Saliency map samples using the baseline method and our method, where the model is tested on different combinations of perturbations shown as columns. 
We show the original image, perturbed image with a chosen effect, saliency map of the baseline model, and saliency map of our method from top to bottom rows.
Using our method, the network focuses more on the important areas (e.g., road in front) instead of random areas on the perturbed images. 
}
\label{fig:salience_map}
\vspace*{-2em}
\end{figure}

\modified{
We also show the t-SNE~\citep{maaten2008visualizing} visualization of feature embeddings from the baseline and proposed method in
Fig.~\ref{Fig:feature_visualization_tsne} of Appendix~\ref{Apd:tsne}. 
Features from our method are more uniformly distributed, indicating the reduction of the domain gaps created by the perturbations, thus improving robustness.}

\begin{table*}[t]
  \centering
  \scalebox{0.8} { 
  \begin{tabular}{c|c|ccc|ccc|ccc}
    \toprule
    & \multicolumn{10}{c}{Scenarios} \\
        \midrule
      & \multicolumn{1}{c}{Clean} & \multicolumn{3}{c}{Single Perturbation} & \multicolumn{3}{c}{Combined Perturbation} & \multicolumn{3}{c}{Unseen Perturbation} \\
    \midrule
  Method& AmAPI$\uparrow$ & MmAPI$\uparrow$ & AmAPI$\uparrow$ & mCE$\downarrow$ & MmAPI$\uparrow$ & AmAPI$\uparrow$ & mCE$\downarrow$ & MmAPI$\uparrow$ & AmAPI$\uparrow$ & mCE$\downarrow$\\
    \midrule
    
Our method & -1.12 & 16.21 & 3.40 & 95.72 & 7.53 & 4.94 & 94.92 & 5.88 & 2.93 & 96.86\\
    \bottomrule
  \end{tabular}}
 \caption{
\modified{Performance improvement of our algorithm for detection task against the baseline performance~\citep{bochkovskiy2020yolov4}. Our method outperforms the baseline in most cases, with about 3\%-5\% mAP improvement on average, while reducing the mCE by 3.14\%-5.18\%. Note: the baseline performance is 0 for AmAPI and MmAPI, and 100 for mCE.}}
  \label{tb:comparison_detection}
\end{table*}

\modified{
\textbf{Performance on detection task:}
We also test our algorithm on the detection task in autonomous driving. We use the Audi dataset~\citep{geyer2020a2d2} (3D Bounding Boxes) and the Yolov4 network~\citep{bochkovskiy2020yolov4} as base settings, and then implement our algorithm based on Yolov4. Table~\ref{tb:comparison_detection} shows that our algorithm also improves the model robustness in most scenarios (about 3\%-5\% mAP improvement on average). 
}

\vspace*{-0.5em}
\subsection{Generalization}
\vspace*{-0.5em}
In this work, we introduce an efficient and effective computational framework, which incorporates sensitivity analysis and a systematic mechanism to improve the performance of a learning algorithm for autonomous driving. The framework performs well on both the original dataset and the simulated adversarial scenarios due to multiple perturbations defined on an influential set of important image attributes. Our method can be easily extended and applied beyond the set of factors and the learning algorithm analyzed in this paper. It can also generalize to analyzing any arbitrarily high number of image/input factors, other learning algorithms, and multimodal sensor data. Lastly, other autonomous systems where the {\em perception-to-control} functionality plays a key role can possibly benefit from our technique as well.

%% file: sections/benchmark.tex
\vspace*{-0.5em}
\subsection{Benchmarking Datasets}
\vspace*{-0.5em}
\label{sec:benchmark}

\modified{
We plan to release our driving datasets with perturbations for benchmarking. The benchmark will contain three collections, including Audi~\citep{geyer2020a2d2}, Honda~\citep{ramanishka2018toward}, and SullyChen dataset~\citep{NVIDIA-Data}. Each of them will contain a base dataset, datasets with five levels of perturbation in blur, noise, and distortion, ten levels of variations in the channels R, G, B, H, S, V, multiple combined perturbations over all nine factors, and five levels of each unseen simulated factor, including snow, fog, frost, motion blur, zoom blur, pixelate, and jpeg compression
using ImageNet-C. There are 360 datasets and about 2.2M images in total. 
The ground-truth steering angles for all images will also be provided for validation, along with the code of perturbed data generation (that can be used to improve the robustness of other learning tasks) and the code of our algorithm.
%
%
We detail the parameters used to generate the datasets in Appendix~\ref{Apd:perturbed-datasets}. 
}

%% file: sections/conclusion.tex
\vspace*{-1em}
\section{Conclusion and Future Work}
\vspace*{-0.5em}

In this paper, we first analyze the influence of different image-quality attributes on the performance of the ``learning to steer'' task for autonomous driving. We have studies three image-level effects (i.e., blur, noise, and distortion) and six channel-level effects (i.e., R, G, B, H, S, V). We observe that image degradations due to these effects can impact task performance at various degrees. By using FID as the unifying metric, we conduct sensitivity analysis in the MA-FID space. Leveraging the sensitivity analysis results, we propose an effective and efficient training method to improve the generalization of learning-based steering under various image perturbations. Our model not only improves the task performance on the original dataset, but also
achieves significant performance improvement on datasets with a mixture of perturbations (up to 48\%), as well as unseen adversarial examples including snow, fog, and frost (up to 29\%). These results show that our model is one of the most general methods for the image-based autonomous driving system. We will release the datasets generated in this work for benchmarking the robustness of learning algorithms for autonomous driving, as well as the code itself.

Our method currently uses discretization to achieve efficient training, but further optimization for our implementation is possible. For example, the efficiency of our technique may be further improved using other methods, such as the reweighting strategy~\citep{ren2018learning}. In summary, we believe that our framework is generalizable to other image factors, learning algorithms, multimodal sensor data, and tasks in other application domains. These are all natural directions worth further exploration.

%% file: sections/appendix.tex
\clearpage
\newpage 

\appendix
\section{Appendix}



\subsection{Dataset Samples}
\label{Apd:image_samples}
\begin{figure*}[h]
\begin{center}
  \includegraphics[width=\linewidth]{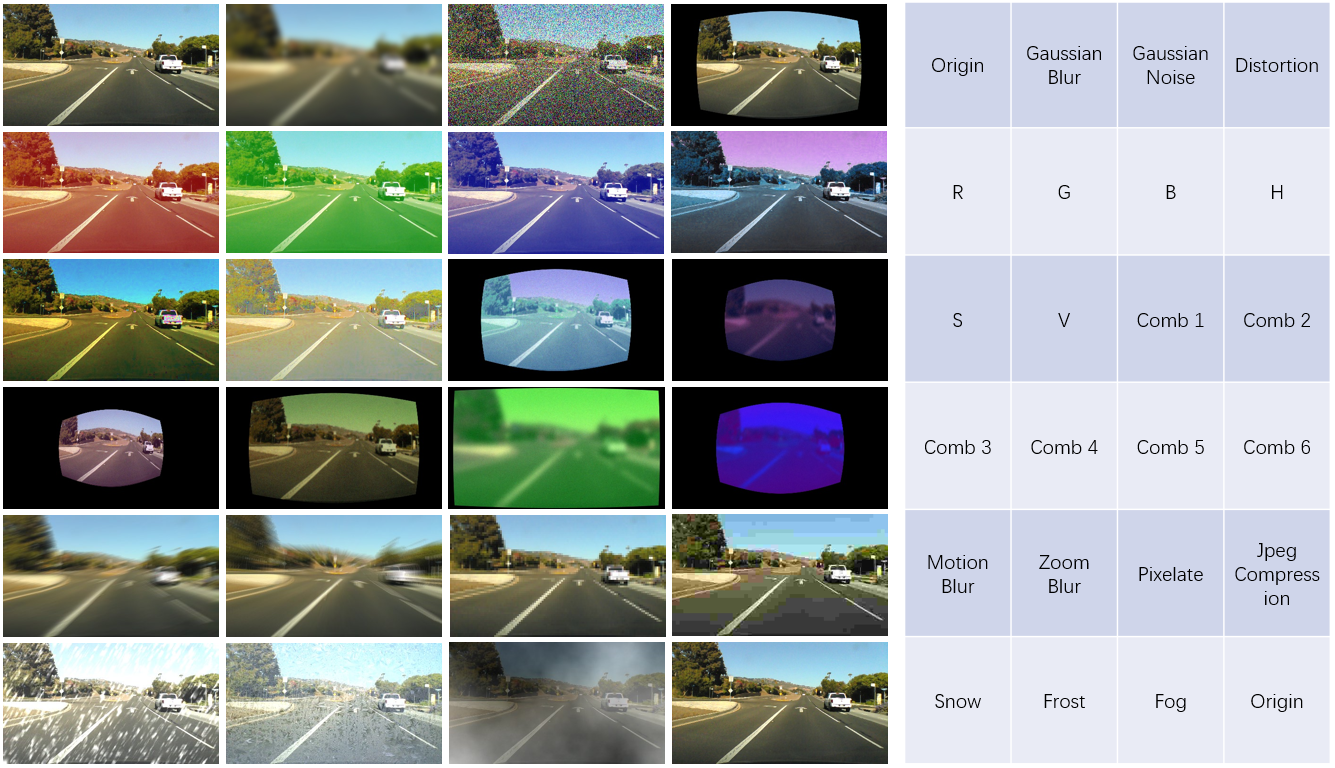}
\end{center}   
\caption{Sample images of our benchmark. We show our benchmark has 22 different types of perturbations. Also, we have 10 levels for R, G, B, H, S, V (5 levels in darker and 5 levels in lighter shades), and 5 levels for each of the other types of perturbations.
}
\label{fig:benchmark_sample}
\end{figure*}

We show different kinds of perturbations in our benchmarks in Fig.\ref{fig:benchmark_sample}. Specifically, our benchmarks include 9 basic types of perturbations, including Gaussian blur, Gaussian noise, radial distortion, and RGB and HSV channels. Another type of datasets include multiple perturbations, where we create multiple random combinations of the basic perturbations. We also include 7 types of previously unseen perturbations (during training) from ImageNet-C, which are snow, fog, frost, motion blur, zoom blur, pixelate, and jpeg compression.  For each type of perturbation, we generate 5 or 10 levels of varying intensity based on sensitivity analysis in the FID-MA space.

\subsection{Perturbed Datasets}
\label{Apd:perturbed-datasets}
In our sensitivity analysis experiments, we first select 10 samples for each of blur, noise, distortion, channel (R, G, B, H, S, V) darker, and channel lighter, then reduce to $n=5$. We set $n=5$ since a smaller number like $n=2$ will decrease the algorithm performance greatly, while a larger number like $n=8$ will decrease the efficiency dramatically.

The final representative datasets from the sensitivity analysis and used for improving the generalization of the learning task are introduced in the following. 
\begin{itemize}
    \item $R$: the base dataset, Audi~\citep{geyer2020a2d2}, Honda~\citep{ramanishka2018toward}, or SullyChen~\citep{NVIDIA-Data} dataset;
    \item $B1, B2, B3, B4, B5$: add Gaussian blur to $R$ with standard deviation $\sigma=1.4$, $\sigma=2.9$, $\sigma=5.9$, $\sigma=10.4$, $\sigma=16.4$, which are equivalent to using the kernel (7, 7), (17, 17), (37, 37), (67, 67), (107, 107), respectively;
    \item $N1, N2, N3, N4, N5$: add Gaussian noise to $R$ with $(\mu=0, \sigma=20)$, $(\mu=0, \sigma=50)$, $(\mu=0, \sigma=100)$, $(\mu=0, \sigma=150)$, $(\mu=0, \sigma=200)$, respectively;
    \item $D1, D2, D3, D4, D5$: distort $R$ with the radial distortion $(k_1=1, k_2=1)$, $(k_1=10, k_2=10)$, $(k_1=50, k_2=50)$, $(k_1=200, k_2=200)$, $(k_1=500, k_2=500)$, respectively. $k_1$ and $k_2$ are radial distortion parameters, the focal length is $1000$, and the principle point is the center of the image.
    \item $RD1/ RL1, RD2/RL2, RD3/RL3, RD4/RL4, $ $RD5/RL5$: modify the red channel of $R$ to darker (D) / lighter (L) values with $\alpha=0.02$, $\alpha=0.2$, $\alpha=0.5$, $\alpha=0.65$, $\alpha=1$.
    \item $GD1/GL1, GD2/GL2, GD3/GL3, GD4/GL4, $ $GD5/GL5$: modify the green channel of $R$ to darker (D) / lighter (L) values with $\alpha=0.02$, $\alpha=0.2$, $\alpha=0.5$, $\alpha=0.65$, $\alpha=1$.
    \item For B, H, S, V channels, we use similar naming conventions for notation as for the red and green channels. 
    \item $Comb1$: $R_\alpha = -0.1180$, $G_\alpha = 0.4343$, $B_\alpha = 0.1445$, $H_\alpha = 0.3040$, $S_\alpha = -0.2600$, $V_\alpha = 0.1816$, $Blur_\sigma = 3$, $Noise_\sigma = 10$, $Distort_k = 17$
    \item $Comb2$: $R_\alpha = 0.0420$, $G_\alpha = -0.5085$, $B_\alpha = 0.3695$, $H_\alpha = -0.0570$, $S_\alpha = -0.1978$, $V_\alpha = -0.4526$, $Blur_\sigma = 27$, $Noise_\sigma = 7$, $Distort_k = 68$
    \item $Comb3$: $R_\alpha = 0.1774$, $G_\alpha = -0.1150$, $B_\alpha = 0.1299$, $H_\alpha = -0.0022$, $S_\alpha = -0.2119$, $V_\alpha = -0.0747$, $Blur_\sigma = 1$, $Noise_\sigma = 6$, $Distort_k = 86$
    \item $Comb4$: $R_\alpha = -0.2599$, $G_\alpha = -0.0166$, $B_\alpha = -0.2702$, $H_\alpha = -0.4273$, $S_\alpha = 0.0238$, $V_\alpha = -0.2321$, $Blur_\sigma = 5$, $Noise_\sigma = 8$, $Distort_k = 8$
    \item $Comb5$: $R_\alpha = -0.2047$, $G_\alpha = 0.0333$, $B_\alpha = 0.3342$, $H_\alpha = -0.4400$, $S_\alpha = 0.2513$, $V_\alpha = 0.0013$, $Blur_\sigma = 35$, $Noise_\sigma = 6$, $Distort_k = 1$
    \item $Comb6$: $R_\alpha = -0.6613$, $G_\alpha = -0.0191$, $B_\alpha = 0.3842$, $H_\alpha = 0.3568$, $S_\alpha = 0.5522$, $V_\alpha = 0.0998$, $Blur_\sigma = 21$, $Noise_\sigma = 3$, $Distort_k = 37$
\end{itemize}

The datasets $Comb1$ through $Comb6$ are generated by randomly sampling parameters of each type of perturbation, e.g. blur, noise, distortion, and RGB and HSV channels, and combining these perturbations together. The parameters listed here are the parameters for the corresponding examples used in the experiment. 

\subsection{Dataset details}
\label{Apd:dataset_details}
\modified{
We use Audi dataset~\citep{geyer2020a2d2}, Honda dataset~\citep{ramanishka2018toward}, and SullyChen dataset~\citep{NVIDIA-Data}. Among the  autonomous driving datasets, Audi dataset is one of the latest dataset (2020), Honda dataset is one of the dataset that has large amout of driving videos (over 100+ videos), and SullyChen dataset is collected specifically for steering task and has a relatively long continuous driving image sequence on a road without branches and has relatively high turning cases.
}

\modified{
For Audi dataset~\citep{geyer2020a2d2}, we use the "Gaimersheim" package which contains about 15,000 images with about 30 FPS. For efficiency, we adopt a similar approach as in \citet{bojarski2016end} by further downsampling the dataset to 15 FPS to reduce similarities between adjacent frames, keep about 7,500 images and align them with steering labels. 
For Honda dataset~\citep{ramanishka2018toward}, which contains more than 100 videos, we first select 30 videos that is most suitable for learning to steer task, then we extract 11,000 images from them at 1 FPS, and align them with the steering labels. 
For SullyChen dataset~\citep{NVIDIA-Data}, images are sampled from videos at 30 frames per second (FPS). We then downsample the dataset to 5 FPS. The resulting dataset contains approximately 10,000 images.
All of them are then randomly splited into training/validation/test data with approximate ratio 20:1:2.}

\modified{There are several good autonomous driving datasets, but not all of them are suitable for the end-to-end learning to steer task. For example, Waymo~\citep{sun2020scalability}, KITTI~\citep{Geiger2013IJRR}, Cityscapes~\citep{cordts2016cityscapes}, OxfordRoboCar~\citep{RobotCarDatasetIJRR}, Raincouver~\citep{tung2017raincouver}, etc., do not contain steering angle labels. Some well-known simulators like CARLA~\citep{CARLA} can generate synthetic dataset, but our work focuses on the real-world driving using real images. There are also several other datasets contain steering angle labels (e.g., nuScenes~\citep{nuscenes2019}, Ford AV~\citep{agarwal2020ford}, Canadian Adverse Driving Conditions~\citep{pitropov2020canadian}, etc), but we didn't use them all because the results on the three datasets we choosed can already show the effectiveness of our method.
}

\subsection{FID-MA and L2D-MA Diff}
\label{Apd:FID_L2D}
\begin{figure*}[t]
\begin{center}
  \includegraphics[width=\textwidth]{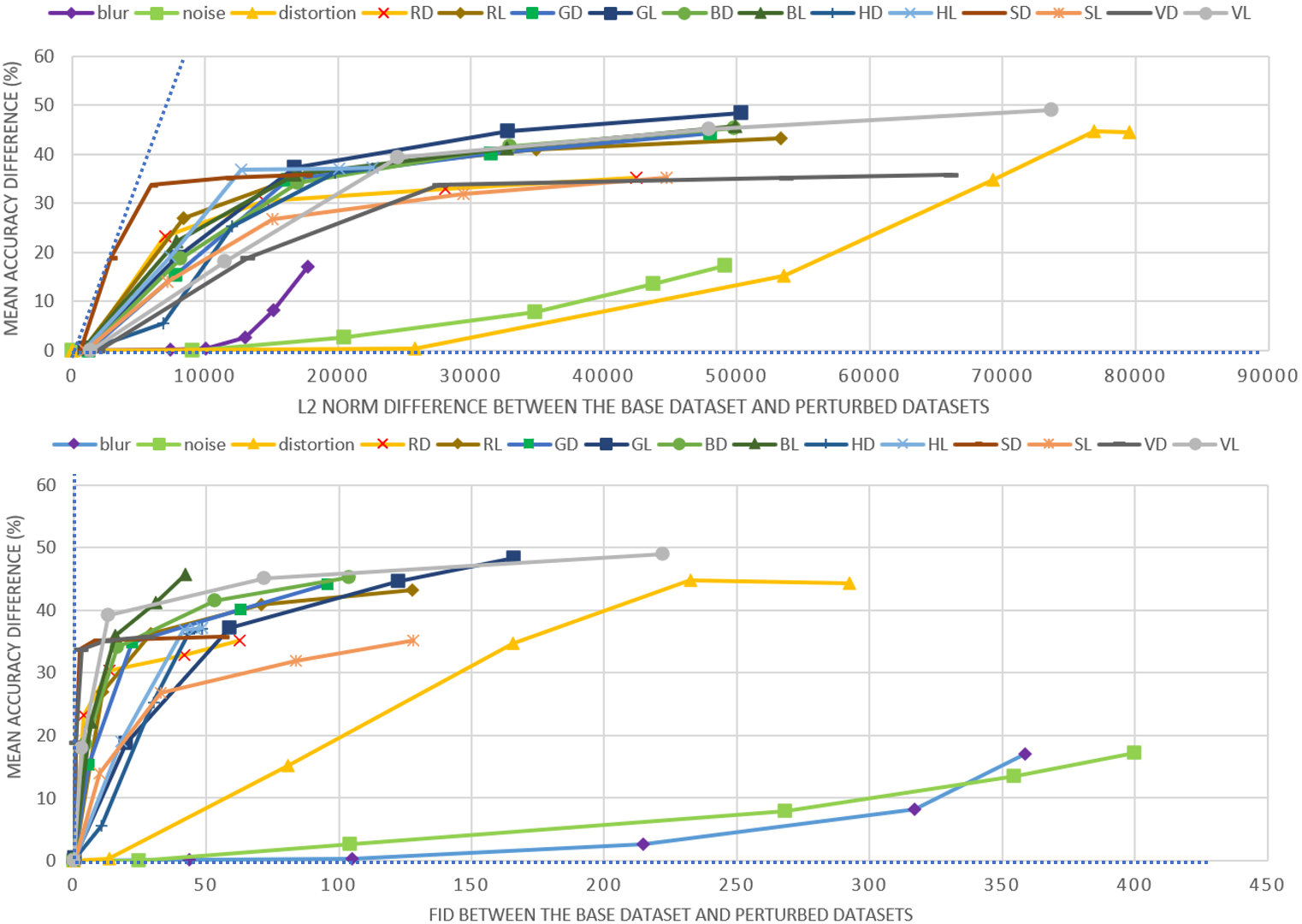}
\end{center}   
\caption{
\modified{The relationship between L2 norm distance and MA difference (top), and the relationship between FID and MA difference (bottom). The FID space can better capture the difference among various factors affecting image quality better than the L2D space, i.e., the range of the curves' first-order derivative is larger in FID space than in L2D space (see the angle between the two dot lines).}}
\label{fig:FID_L2D_MA_large}
\end{figure*}

We illustrate the relationship between FID and Mean Accuracy (MA) Difference, and the relationship between L2 norm distance (L2D) and Mean Accuracy (MA) Difference in Figure~\ref{fig:FID_L2D_MA_large}. As shown in the figure, the FID space can better capture the difference among various factors affecting image quality better than the L2D space, i.e., the range of the curves' first-order derivative is larger in FID space than in L2D space (see the angle between the two dot lines).

\subsection{t-SNE visualization}
\label{Apd:tsne}
\begin{figure*}[t]
\begin{center}
  \includegraphics[width=\linewidth]{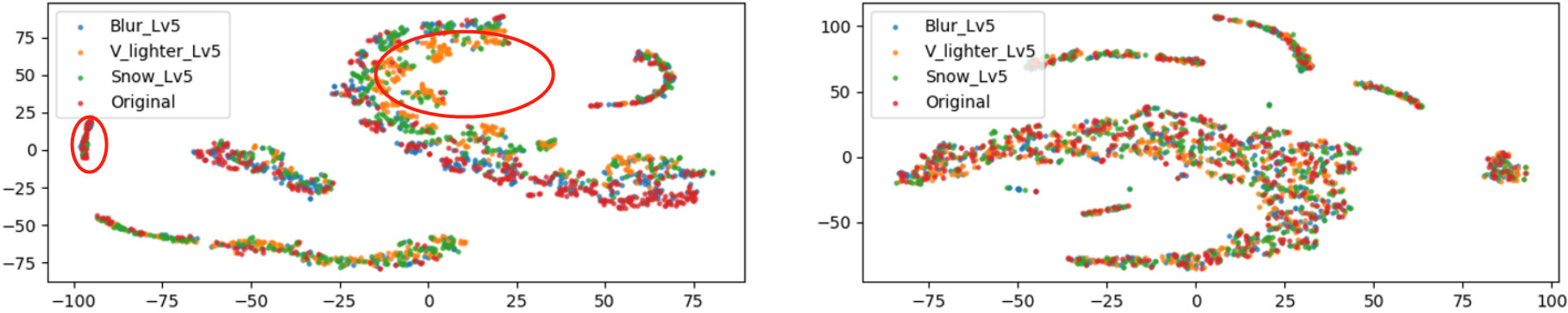}
\end{center}   
\caption{
\modified{t-SNE~\citep{maaten2008visualizing} visualization for features achieved from networks trained by baseline method (left) and our method (right). The features from baseline method are more clustered by color (e.g., the left circle in the left image mainly contains red dots, and the right circle in the left image mainly contains yellow dots), indicating there are domain gaps between the perturbed data and original data, while the features from our method are more uniformly distributed, suggesting that our method is able to reduce the domain gaps due to perturbations, i.e., improving the robustness.}}
\label{Fig:feature_visualization_tsne}
\end{figure*}

\modified{
We show the t-SNE~\citep{maaten2008visualizing} visualization of feature embedings for baseline method and our method in Fig.~\ref{Fig:feature_visualization_tsne}. The features from baseline method are more clustered by color (e.g., the left circle in the left image mainly contains red dots, and the right circle in the left image mainly contains yellow dots), indicating there are domain gaps between the perturbed data and original data; while the features from our method are more uniformly distributed, suggesting that our method is able to reduce the domain gaps from perturbations, i.e., improve the robustness.
}

\subsection{Experiment data}
\label{Apd:experiment_data}



To quantify our results, we collected mean accuracy (MA) measurements from each experiment, for each pairwise factor and level across methods. Table ~\ref{tb:quality_image} shows the mean accuracy measurements for blur, noise, and distortion factors. The same is of table ~\ref{tb:quality_channel}, where mean accuracy is measured across levels of RGB or HSV color channels, where each channel serves as a single corruption factor. Table ~\ref{tab:baseline_combo} presents the MA measurements for scenarios with a combination of factors, and Table~\ref{tab:baseline_unseen} presents the MA measurements for scenes with previously unseen factors. 

\begin{figure*}[h]
\begin{center}
  \includegraphics[width=\linewidth]{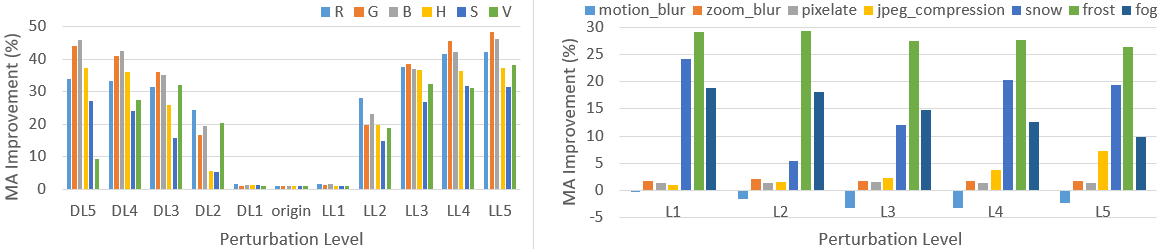}
\end{center}   
\caption{MA improvement with our method compared to the baseline. Our method achieve great improvement on extreme cases for channel-level factors and unseen weather conditions, e.g., up to {\bf 48.9\%} MA improvement in the Lighter Level 5 (LL5) of the Green channel in the left figure, or up to {\bf 29.4\%} MA improvement in the Level 2 (L2) of frost effect among unseen test data in the right figure.}
\label{fig:MA_channel_unseen_improvement}
\end{figure*}

\modified{Fig.~\ref{fig:MA_channel_unseen_improvement} shows the MA improvement with our method compared to the baseline. Our method achieve great improvement on extreme cases for channel-level factors and unseen weather conditions.}

\begin{table*}[h]
  \caption{Mean Accuracy of training  (in \%) using the baseline model and ours, tested on datasets with different levels of blur, noise, and distortion. Levels range from L1 to L5. We achieve up to 10.5\% in performance gain (see bold number pair).}
  \label{tb:quality_image}
  \centering
  
  \scalebox{1.0}{
  \begin{tabular}{c|c|ccccc}
    \toprule
Method & Factor & L1 & L2 & L3 & L4 & L5\\
    \midrule
         &  blur  & 88.2 & 88.1 & 86.1 & 81.2 & 73.3 \\
        baseline & noise & 88.3 & 86.0 & 81.4 & 76.4 & 73.2 \\
         & distortion  & 88.6 & \textbf{75.0} & 57.7 & 48.8 & 49.2 \\
        \midrule 
         & blur  & 89.2 & 89.5 & 88.8 & 82.4 & 75.5 \\
        ours & noise &  89.1 & 88.7 & 88.5 & 85.5 & 82.7 \\
         & distortion & 89.1 & \textbf{85.5} & 63.1 & 56.5 & 50.6 \\
    \bottomrule
  \end{tabular}
  }
\end{table*}

\begin{table*}[h]
  \caption{Mean accuracy (MA) of training (in \%) using the baseline model and ours, tested on datasets with different levels of R, G, B, and H, S, V channel values. DL denotes "darker level", which indicates a level in the darker direction of the channel, while LL indicates "lighter level", which indicates the lighter direction, on levels 1 to 5. We achieve up to {\bf 48.9\%} in performance gain (see bold number pair).}
  \label{tb:quality_channel}
  \centering
  \begin{tabular}{c|c|cccccccccc}
    \toprule
 Method & Factor & DL5 & DL4 & DL3 & DL2 & DL1 & LL1 & LL2 & LL3 & LL4 & LL5\\
    \midrule
        & R & 53.2 & 55.4 & 57.9 & 65.1 & 87.8 & 87.7 & 61.4 & 52.1 & 47.4 & 45.1 \\
        & G & 44.2 & 48.2 & 53.5 & 73.0 & 88.5 & 87.9 & 69.6 & 51.2 & 43.7 & \textbf{40.0} \\
        baseline & B & 43.0 & 46.8 & 54.3 & 69.7 & 88.2 & 87.7 & 66.2 & 52.5 & 47.1 & 42.6 \\
        & H & 51.3 & 52.1 & 63.1 & 82.8 & 88.1 & 88.2 & 69.3 & 51.5 & 51.3 & 51.2 \\
        & S & 58.4 & 63.8 & 72.6 & 83.9 & 88.1 & 88.3 & 74.5 & 61.6 & 56.5 & 53.2 \\
        & V & 52.6 & 53.2 & 54.6 & 69.4 & 88.5 & 88.4 & 70.4 & 49.1 & 43.2 & 39.4 \\
        \midrule 
        & R & 87.3 & 88.8 & 89.4 & 89.5 & 89.4 & 89.4 & 89.4 & 89.7 & 89.1 & 87.4 \\
        & G & 88.4 & 89.3 & 89.7 & 89.6 & 89.4 & 89.3 & 89.4 & 89.5 & 89.3 & \textbf{88.9} \\
        ours & B & 89.0 & 89.5 & 89.5 & 89.2 & 89.4 & 89.3 & 89.4 & 89.5 & 89.3 & 88.9 \\
        & H & 88.7 & 88.4 & 89.1 & 88.5 & 89.2 & 89.2 & 89.1 & 88.4 & 87.8 & 88.7 \\
        & S & 85.7 & 87.8 & 88.2 & 89.0 & 89.3 & 89.3 & 89.3 & 88.5 & 88.2 & 84.5 \\
        & V & 61.9 & 80.6 & 86.8 & 89.7 & 89.3 & 89.3 & 89.1 & 81.4 & 74.5 & 77.7 \\
    \bottomrule
  \end{tabular}
\end{table*}

\begin{table*}[h]
    \centering
    \caption{Mean accuracy (MA) of training (in \%) using the baseline model and ours, tested on datasets with several perturbations combined together, including blur, noise, distortion, RGB, and HSV. We achieve up to {\bf 33.3\%} in performance gain (see bold number pair).}
    \label{tab:baseline_combo}
    \scalebox{1.0}{
    \begin{tabular}{c|cccccc}
        \toprule
        Method & Comb1 & Comb2 & Comb3 & Comb4 & Comb5 & Comb6\\
        \midrule 
        baseline & 59.7 & 54.0 & 40.9 & \textbf{50.0} & 54.0 & 56.3 \\
        ours & 71.3 & 61.1 & 65.6 & \textbf{83.3} & 85.6 & 54.5 \\
        \bottomrule
    \end{tabular}
    }
\end{table*}

\begin{table*}[h]
    \centering
    \caption{Mean accuracy (MA) of training (in \%) using the baseline model and ours, tested on datasets with previously unseen perturbations at 5 different levels. These types of unseen perturbations do not appear in the training data, and include motion blur, zoom blur, pixelate, jpeg compression loss, snow, frost, and fog, on intensity levels L1 to L5. We achieve up to {\bf 29.4\%} in performance gain (see bold number pair).}
    \label{tab:baseline_unseen}
    \scalebox{1.0}{
    \begin{tabular}{c|c|ccccc}
    \toprule
        Method & Unseen Factors & L1 & L2 & L3 & L4 & L5 \\
        \midrule
        & motion\_blur & 76.4 & 69.7 & 62.6 & 61.1 & 60.3 \\
        & zoom\_blur & 85.6 & 83.7 & 81.8 & 80.0 & 78.2 \\
        & pixelate & 88.2 & 88.2 & 88.0 & 88.3 & 88.1 \\
        baseline & jpeg\_comp & 88.4 & 88.0 & 87.4 & 85.4 & 82.2 \\
        & snow & 62.8 & 50.7 & 54.9 & 55.5 & 55.3 \\
        & frost & 55.8 & \textbf{52.1} & 51.7 & 51.7 & 51.2\\
        & fog & 58.7 & 55.0 & 52.4 & 50.8 & 48.1 \\
        \midrule 
        & motion\_blur & 76.0  & 68.1 & 59.4 & 57.9 & 58.1\\
        & zoom\_blur & 87.4 & 85.8 & 83.6 & 81.8 & 79.9 \\
        & pixelate & 89.6 & 89.7 & 89.6 & 89.6 & 89.5 \\
        ours & jpeg\_comp & 89.5 & 89.5 & 89.6 & 89.2 & 89.4 \\
        & snow & 86.9 & 56.2 & 66.9 & 75.8 & 74.6 \\
        & frost & 84.9 & \textbf{81.5} & 79.2 & 79.3 & 77.6\\
        & fog & 77.6 & 73.2 & 67.2 & 63.4 & 57.9\\
        \bottomrule
    \end{tabular}
    }
\end{table*}

